\documentclass[runningheads]{llncs}
\usepackage[T1]{fontenc}
\usepackage{graphicx}
\usepackage{algorithm}
\usepackage{algorithmic}
\usepackage{algorithmic}
\usepackage{bm}
\usepackage{multirow}
\usepackage{textcomp}
\usepackage{stfloats}
\usepackage{url}
\usepackage{verbatim}
\usepackage{graphicx}
\usepackage{amsmath,amsfonts}

\begin{document}
\title{Improving Handwritten Text Recognition via 3D Attention and Multi-Scale Training}
\titlerunning{Abbreviated paper title}
%

\author{Zi-Rui Wang}

%


%
\institute{ 
}

%
\maketitle              
\begin{abstract}
The segmentation-free research efforts for addressing handwritten text recognition can be divided into three categories: connectionist temporal classification (CTC), hidden Markov model and encoder–decoder methods. In this paper, inspired by the above three modeling methods, we propose a new recognition network by using a novel three-dimensional (3D) attention module and global-local context information. Based on the feature maps of the last convolutional layer, a series of 3D blocks with different resolutions are split. Then, these 3D blocks are fed into the 3D attention module to generate sequential visual features. Finally, by fusing the visual features and the corresponding global-local context features, a well-designed representation can be obtained. Main canonical neural units including attention mechanisms, fully-connected layers, recurrent units and convolutional layers are efficiently organized into a network and can be jointly trained by the CTC loss and the cross-entropy loss. Experiments on the latest Chinese handwritten text datasets (the SCUT-HCCDoc and the SCUT-EPT) and one English handwritten text dataset (the IAM) show that the proposed method can achieve comparable results with the state-of-the-art methods. The code is available at https://github.com/Wukong90/3DAttention-MultiScaleTraining-for-HTR.

\keywords{Handwritten text recognition  \and Segmentation-free recognition \and 3D attention module \and  Multi-scale training \and Global-local context information.}

\end{abstract}
\section{Introduction}

\begin{figure}[!t]
	\centering
	\includegraphics[width=4.8in,height=2.5in]{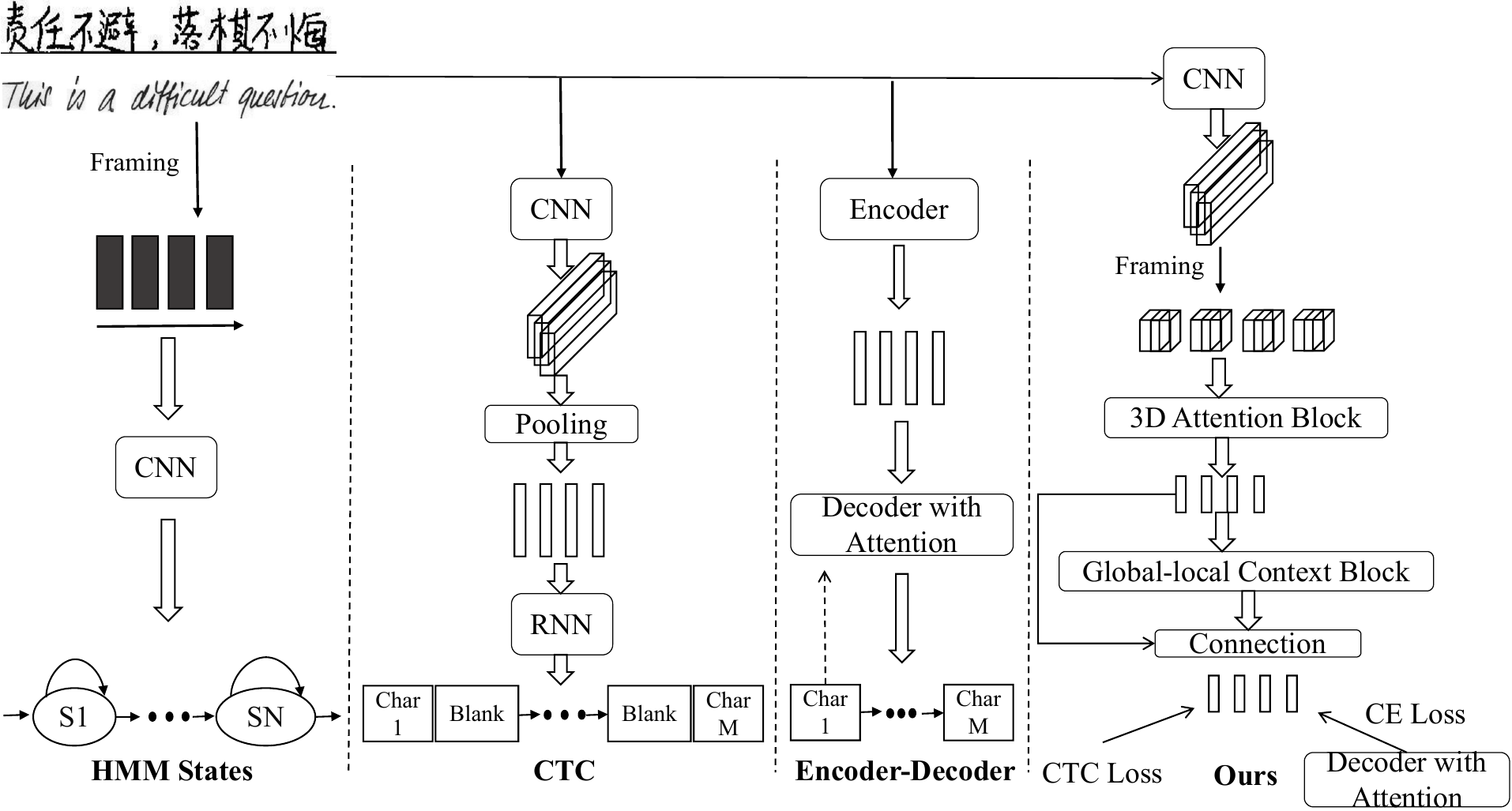}
	\caption{The typical segmentation-free methods and the proposed network. The abbreviations CNN, RNN, Char, S denote convolutional neural network, recurrent neural network, character and state, respectively.}
	\label{difses}
\end{figure}

In text recognition tasks, there are three typical segmentation-free methods, i.e., hidden Markov model (HMM) \cite{espana2010improving,wang2020writer}, connectionist temporal classification (CTC)\cite{graves2006connectionist,messina2015segmentation,shi2016end,dutta2018improving,hoang2021lodenet,wang2022fast} and encoder–decoder (ED) framework \cite{bahdanau2014neural,chowdhury2018efficient,zhang2020radical,ngo2021recurrent}. As shown in Fig.~\ref{difses}, in the HMM-based method, each character is modeled by an HMM and a text line can be represented by cascaded HMMs. A series of frames extracted from an original image by a left-to-right sliding window are assigned to the underlying states. Then, a neural network is used to estimate the posterior probabilities of the states, while the outputs of networks in CTC and ED-based approaches are character classes. In the CTC loss, a special character "blank" and a sophisticated rule are designed to split different characters, and the forward-backward algorithm can efficiently compute the probability of the underlying character sequence. In an encoder-decoder network, an image is usually fed into the encoder to generate the corresponding middle features. Based on the middle representations, the decoder is used to locate and predict the character sequence via attention mechanisms. The common cross-entropy loss can be directly used to adjust the parameters of the ED network. 

It is reasonable to explicitly model the 2D information of characters. However, it is very difficult to expand 1D HMM to 2D HMM \cite{ma2021open} due to the computational complexity. Even for the CTC, ED-based approaches \cite{graves2008offline,puigcerver2017multidimensional,dutta2018improving,chowdhury2018efficient}, early networks usually simply depend on the local receptive field of convolutional layers or recurrent units and gradually shrink the height of feature maps to small pixels via stacked pooling layers, which may lead to information loss. Although some attempts \cite{cheng2017focusing,lee2020recognizing,zhang2020radical,yu2023chinese} explicitly or implicitly use 2D information in scene text recognition and character recognition, there is no obvious evidence showing that these methods can be directly used in long handwritten text recognition. Obviously, the context information is important for the text recognition. In \cite{qiao2020seed,wang2021two,fang2021read}, the visual features and the linguistic knowledge are simultaneously integrated into a network. Actually, based on the visual features extracted from a certain image, the corresponding context information can also be naturally obtained by using recurrent units \cite{shi2016end,cheng2017focusing,shi2018aster,wang2022fast} or  transformers \cite{lee2020recognizing,coquenet2023dan,tan2022pure}.

For the handwritten text recognition task, recently, Hoang et al. \cite{hoang2021lodenet} combined the radical-level CTC loss and the character-level loss while Ngo et al. \cite{ngo2021recurrent} constructed a joint decoder to integrate the visual feature and the linguistic context feature. More recently, Peng et al. \cite{peng2022recognition} built a full convolution network to simultaneously achieve the purpose of the character location, the character bounding boxes prediction and the character prediction. Lin et al. \cite{lin2023building} propose a mobile text recognizer via searching the lightweight neural units. Furthermore, Peng et al. \cite{peng2022pagenet} detect and recognize characters in page-level handwritten text while Coquenet et al. \cite{coquenet2023dan} use a fully convolutional network as the encoder with a transformer decoder for document recognition.

In this paper, inspired by the above three modeling ways, we propose a new recognition network by using a novel 3D attention module and global-local context information. The 3D attention module is employed to explicitly extract 2D information of block features with different resolutions. In detail, the 3D attention module is ingeniously decoupled into a 2D self-attention operation and a 1D attention-based aggregation operation. Based on the outputs (visual features) of the 3D attention module, we further extract the corresponding global-local context information via the self-attention and the recurrent unit, respectively. Finally, by fusing the visual features and the corresponding global-local context features, a well-designed representation for the recognition task can be obtained. In summary, the main contributions of this paper are as follows:
\begin{enumerate}
	\item{Inspired by the typical segmentation-free approaches, we improve the text recognition network by using a novel 3D attention module and global-local context information.}
	\item{Main canonical neural units including attention mechanisms, fully-connected layers, recurrent units and convolutional layers are ingeniously organized into a network.}
	\item{We propose the multi-scale training approach that includes extracting 3D blocks with different resolutions and simultaneously adopting the CTC and the CE losses.}
	\item{Compared with the state-of-the-art methods, the proposed network can achieve comparable results on all datasets. We conduct a comprehensive analysis to verify the effects of the 3D attention module and the different features.}
\end{enumerate}

The remainder of this paper is organized as follows: Section \ref{Me} elaborates on the details of the proposed method. Section \ref{exp} reports the experimental results and analysis. Finally, we discuss the advantages of the proposed method and conclude the paper.

\section{Methodology}
\label{Me}
\begin{figure*}[!t]
	\centering
	\includegraphics[width=4.8in,height=3.0in]{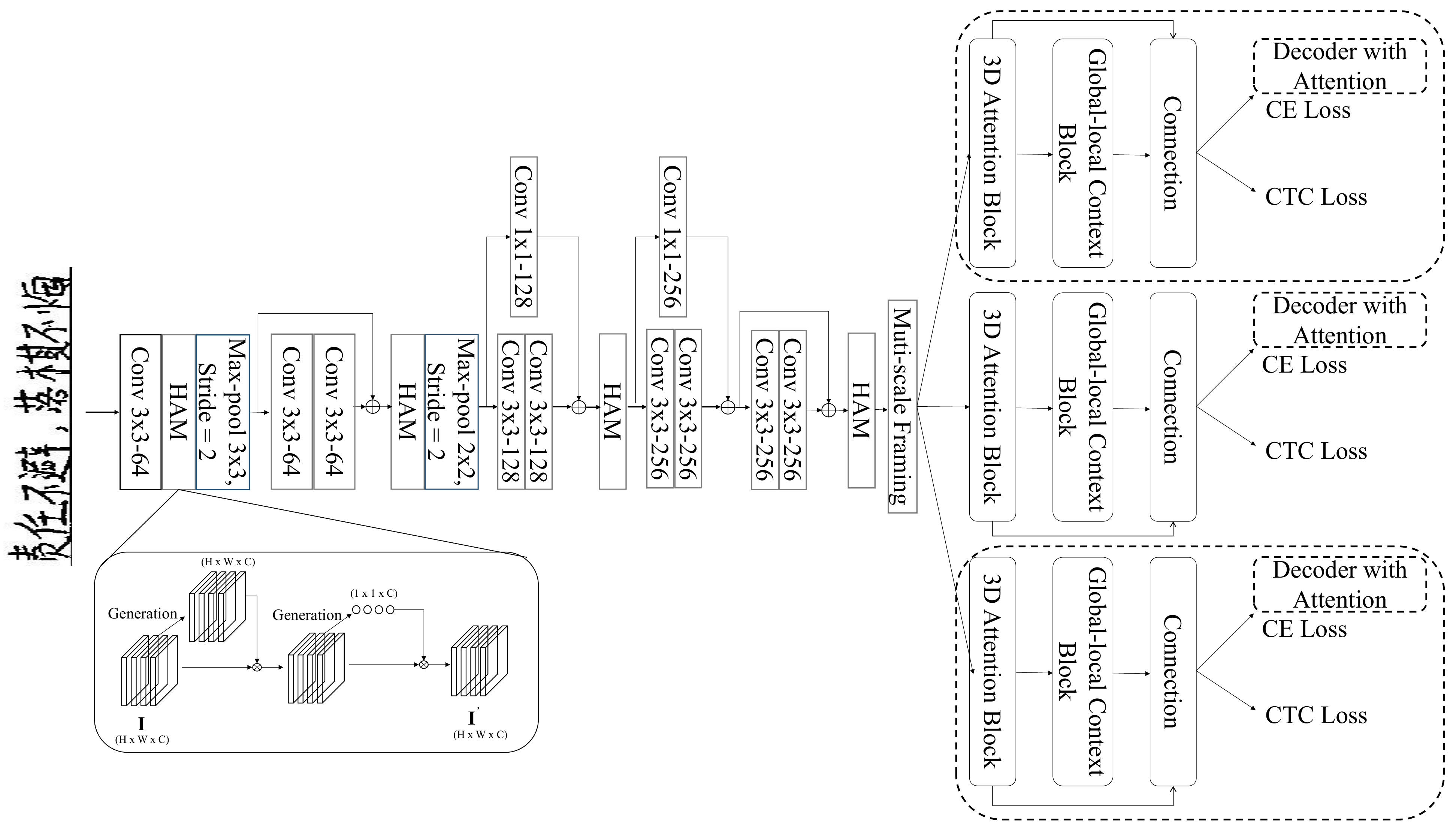}
	\caption{The parts within the dashed lines are only used during the training stage. The details of the 3D attention, global-local context, decoder with attention, CE loss and CTC loss are illustrated in the following sections.}
	\label{nn}
\end{figure*}

\noindent As shown in Fig.~\ref{nn}, the proposed network includes three parts, i.e., the convolutional neural network (CNN), the 3D attention module and the features fusion block. In the CNN, the hybrid attention module (HAM) \cite{wang2022fast} is used and each convolutional layer is equipped with the batch normalization \cite{ioffe2015batch}. For the outputs of the CNN, we use a multi-scale framing strategy to extract different solutions. In this section, we elaborate on the details of the 3D attention module and the global-local context information. Moreover, the multi-scale training pipeline is also shown.

\subsection{3D Attention}
\begin{figure}[!t]
	\centering
	\includegraphics[width=3.5in,height=2.0in]{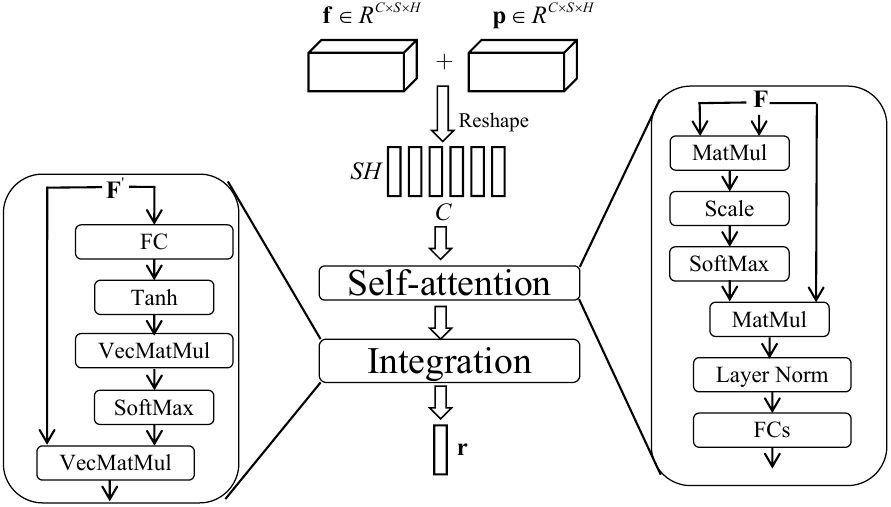}
	\caption{The proposed 3D attention block.}
	\label{3dat}
\end{figure}

\noindent As shown in Fig.~\ref{3dat}, assuming the front CNN has the output tensor ${{\bf{O}} \in {R^{C \times W \times H}}}$ with each feature map ${\bf{O}}_c \in {R^{W \times H}}$, each 3D block represented by a sliding window ${{\bf{f}} \in {R^{C \times S \times H}}}$ from left to right, with the window shift of $S$ pixels, is scanned across the feature maps. Firstly, each location $p$ in the plane $S \times H$ is added the corresponding positional encoding:
\begin{equation}
	{\bf{f}}_{p}[0:C/2] = {\bf{f}}_{p}[0:C/2] + {{\bf{p}}_{\rm{w}}}
\end{equation}
\begin{equation}
	{\bf{f}}_{p}[C/2{\rm{:(C - 1)]}} = {\bf{f}}_{p}[C/2{\rm{:(C - 1)]}} +  {{\bf{p}}_{\rm{h}}}
\end{equation}
where ${{\bf{p}}_{\rm{w}}}$ and ${{\bf{p}}_{\rm{h}}}$ are sinusoidal positional encoding over height and width, respectively, as defined in \cite{vaswani2017attention}:
\begin{equation}
	{{\bf{p}}_{p,2i}} = \sin (p{e^{2i( - \ln (10000)/C)}})
\end{equation}
\begin{equation}
	{{\bf{p}}_{p,2i + 1}} = \cos (p{e^{2i( - \ln (10000)/C)}})
\end{equation}
here $i$ is the index along hidden dimensions. And then, we reshape each 3D block from ${C \times S \times H}$ to $SH \times C$ and conduct a self-attention operation for these sequential vectors ${\bf{f}}_{p}$ as Eq.\ref{seat}:

\begin{equation}
	{{\bf{F}}^{'}} = {\rm{softmax(}}\frac{{{\bf{F}}{{\bf{F}}^T}}}{s}{\rm{)}}\bf{F} \label{seat}
\end{equation}
The matrices ${\bf{F}}$, ${\bf{F}}^{'}$ pack all vectors ${\bf{f}}_p$ and the corresponding ${\bf{f}}^{'}_p$,respectively. Finally, a representation vector $\bf{r}$ for all ${\bf{f}}_p^{'}$ can be obtained:
\begin{equation}
	{e_p} = {{\bf{w}}^{\rm{T}}} \tanh {({\bf{W}}{{\bf{f}}_p^{'}}+\bf{b})}  
\label{at1}
\end{equation}
\begin{equation}
	{\alpha _p} = \frac{{{e_p}}}{{\sum\limits_i {{e_i}} }} 
\label{at2}
\end{equation}
\begin{equation}
	{\bf{r}} = \sum\limits_p {{\alpha _p}{\bf{f}}_p^{'}} 
\label{at3}
\end{equation}
Similar to the transformer encoder, the vectors ${\bf{f}}_p^{'}$ are fed into a layer normalization (Layer Norm) followed by two fully connected layers (FCs) before computing Eq.\ref{at1}-\ref{at3}.

\subsection{Features Fusion}
\begin{figure*}[!t]
	\centering
	\includegraphics[width=4.8in,height=2.0in]{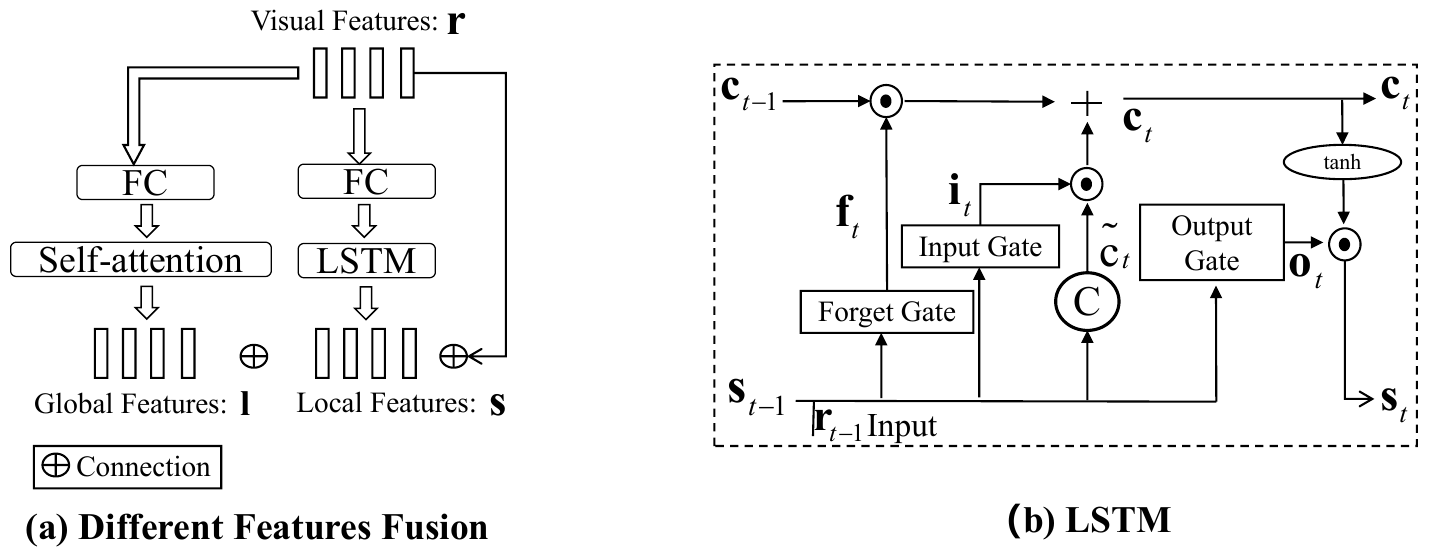}
	\caption{The different features (visual features, global-local features) are combined before the classification layer and the details of the LSTM are illustrated.}
	\label{feain}
\end{figure*}

\noindent Through the 3D attention module, a series of 3D feature maps can be transformed into the corresponding visual features ${\bf{r}}_t (t=0...T)$. As shown in Fig.~\ref{feain}-(a), for the visual feature ${\bf{r}}_t$, the global context feature ${\bf{l}}_t$ and the local context ${\bf{s}}_t$ can be extracted by using the self-attention mechanism and the recurrent units, respectively. We directly use the computation method of the ${\bf{f}}^{'}$ in Fig.~\ref{3dat} to obtain the global context information ${\bf{l}}_t$. For the local context features, the standard LSTM (Fig.~\ref{feain}-(b)) is used. 
${\bf{r}}_t$ represents the input at time $t$ and ${\bf{s}}_t$ is the corresponding output, $\bf{i}$, $\bf{g}$, $\bf{o}$ and $\bf{c}$ are the outputs of input gate, forget gate, output gate and cell vectors, respectively. Finally, the visual feature ${\bf{r}}_t$, the global feature ${\bf{l}}_t$ and the local feature ${\bf{s}}_t$ are simply contacted together:
\begin{equation}
	{{\bf{v}}_t} = {{\bf{r}}_t} \oplus {{\bf{l}}_t} \oplus {{\bf{s}}_t}
\end{equation}

The well-defined representations $\bf{v}$ can be directly fed into a classification layer and also be used as the input of the decoder during the joint training of the CTC and the CE losses.

\subsection{Multi-Scale Training}
\noindent The proposed multi-scale training includes two parts. Firstly, as shown in Fig.~\ref{nn}, in the training stage, multiple feature sequences with different frame lengths are extracted. Although parallel branches including the same 3D attention block and global-local context block are used, we only retained the corresponding branch of the frame length 3 during the inference stage in our experiments.

Secondly, we use the joint training of the CTC and the CE losses. Given the feature sequence $\bf{V}$ of a text line image, the text recognition task is to find the corresponding underlying n-character sequence ${\bf{C}} = \{C_1, C_2, ... , C_n\}$, i.e, compute the posterior probability $p(\bf{C}|\bf{V})$. The CTC can be regarded as a loss function of neural networks:
\begin{equation}
	{L_{{\rm{CTC}}}}({\bf{\Theta}}) = - \log (\sum\limits_{{\bm{\pi }}:\varphi ({\bm{\pi }}) = {\bf{C}}} {p({\bm{\pi }}|{\bf{V}},{\bf{\Theta }}))} \label{ctc_loss}
\end{equation}

\begin{figure}
	\centering
	\includegraphics[width=2.3in,height=2.5in]{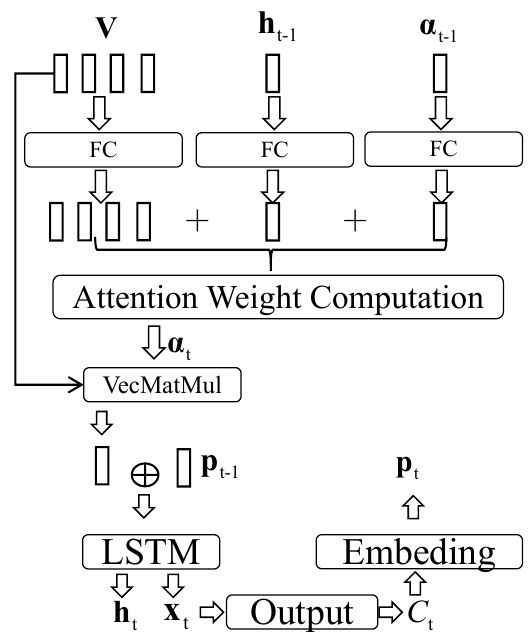}
	\caption{The decoder with attention. The Attention Weight Computation module is similar to the Integration module in Fig.~\ref{3dat}}
	\label{dec}
\end{figure}

\noindent where $\bf{\Theta}$ represents the parameters of the recognition network and $\bm{\pi}$ is the predicted character sequence under the constraint of $\varphi ({\bm{\pi}}) = {\bf{C}}$. The function $\varphi$ only retains one of the consecutive adjacent repeating characters and removes the ‘blank’ characters. The CTC loss (Eq.\ref{ctc_loss}) can be efficiently computed by using the forward-backward algorithm \cite{graves2006connectionist}. For the auxiliary decoder in Fig.~\ref{dec}, the CE loss directly predicts the probability $p(C_{t}|{\bf{x}}_t)$:
\begin{equation}
	{L_{{\rm{CE}}}}({\bf{\Theta,\Gamma}}) = \prod\limits_t {p({C_t}|{{\bf{x}}_t})}
\end{equation}

\noindent $\bf{\Gamma}$ is the parameters of the decoder. Based on the feature sequence $\bf{V}$, the vector ${{\bf{x}}_t}$ at the time step $t$ can be computed using the following attention method:
\begin{equation}
	{{\bf{q}}_t} = \tanh ({{\bf{W}}_{\rm{q}}}{{\bf{h}}_{t - 1}} + {{\bf{b}}_{\rm{q}}})
\end{equation}
\begin{equation}
	{{\bf{k}}_i} = \tanh ({{\bf{W}}_{\rm{k}}}{{\bf{v}}_i} + {{\bf{b}}_{\rm{k}}})
\end{equation}
\begin{equation}
	{{\bf{a}}_t} = \tanh ({{\bf{W}}_{\rm{a}}}{{\bm{\alpha}}_{t - 1}})
\end{equation}
\begin{equation}
	{e_{t,i}} = {{\bf{w}}^{\rm{T}}}\tanh ({{\bf{q}}_t} + {{\bf{k}}_i} + {{\bf{a}}_t})
\end{equation}
\begin{equation}
	{\alpha _{t,i}} = \frac{{{e_{t,i}}}}{{\sum\limits_i {{e_{t,i}}} }}
\end{equation}
\begin{equation}
	{{\bf{x}}_t} = {\rm{LSTM}}((\sum\limits_i {{\alpha _{t,i}}{{\bf{v}}_i}} ) \oplus {{\bf{p}}_{t - 1}})
\end{equation}
where ${\bf{h}}_{t - 1}$,${{\bf{p}}_{t - 1}}$are the hidden state of the LSTM and the decoded embedding vector at the time step $t-1$, respectively.

Finally, Alg.\ref{alg:trp} describes the joint training of the CTC and the CE Losses.
\begin{algorithm}[htb]
	\caption{The joint training pipeline of the CTC and the CE losses.}
	\label{alg:trp}
	\begin{algorithmic}[1]
		\REQUIRE ~~\\
		The randomly initialized parameter sets \{$\bf{\Theta}$, $\bf{\Gamma}$\}; \\ 
		The loss functions $L_{\rm{CTC}}$, $L_{\rm{CE}}$ and the coressponding weight coefficients ${\lambda _1,\lambda _2}$  in the training stage of the main recognition network (MRN($\bf{\Theta}$)), and the auxiliary decoder (Dec($\bf{\Gamma}$)), respectively.
		\STATE Optimize the MRN parameter set $\bf{\Theta}$ by using the Adam algorithm \cite{kingma2014adam}.\\
		$\bf{\Theta} = \rm{Adam}({\bf{\Theta}},L_{\rm{CTC}})$
		\STATE Jointly train the parameter sets ${\bf{\Theta}},{\bf{\Gamma}}$ based on the CTC and the CE losses.\\
		$L = {\lambda _1}L_{\rm{CTC}} + {\lambda _2}L_{\rm{CE}}$ \\
		$\bf{\Theta,\Gamma} = \rm{Adam}({\bf{\Theta,\Gamma}},L)$
		\RETURN The MRN($\bf{\Theta}$).
	\end{algorithmic}
\end{algorithm}

\section{Experiments}
\label{exp}
\subsection{Datasets and Evaluation Metrics}

\begin{table}
	\begin{center}
		\caption{The number of text lines, characters and classes in the SCUT-HCCDoc and the SCUT-EPT.}
		\label{Stanum}
		\begin{tabular}{|c|cc|cc|}
			\hline
			\multirow{2}{*}{Type}      & \multicolumn{2}{c|}{SCUT-HCCDoc}                                                           & \multicolumn{2}{c|}{SCUT-EPT}                                                      \\ \cline{2-5} 
			& \multicolumn{1}{c|}{Training set}  &Test set & \multicolumn{1}{c|}{Training set} & Test set \\ \hline
			Text lines                  & \multicolumn{1}{c|}{93,254}                      &23,484       & \multicolumn{1}{c|}{40,000 }                     &10,000      \\ \hline
			Characters & \multicolumn{1}{c|}{925,200}                       & 230,019       &  \multicolumn{1}{c|}{1,018,432}                       & 248,730    \\ \hline
			Classes & \multicolumn{1}{c|}{5,922}                   & 4,435      &  \multicolumn{1}{c|}{4,058}                    &    3,236   \\
			\hline
		\end{tabular}
	\end{center}
\end{table}

\noindent The proposed network is validated on two Chinese handwritten text datasets(the SCUT-HCCDoc\cite{zhang2020scut} and the SCUT-EPT\cite{zhu2018scut}) and one English handwritten text dataset: the IAM\cite{marti2002iam}.

All the images of SCUT-HCCDoc were obtained by Internet search. According to certain rules, only 12,253 images were preserved from the initial 100 thousand candidate images. These images were randomly split into the training and test sets with a ratio of 4:1. After splitting, the training set contains 9,801 images with 93,254 text instances and 925,200 characters. The test set contains 2,452 images with 23,484 text instances and 230,019 characters.

The Chinese handwritten text data set SCUT-EPT contains 4,250 categories of Chinese characters and symbols, totaling 1,267,162 characters, and it was split into 40,000 training text lines and 10,000 test text lines. These samples were collected from examination papers and were written by 2,986 students, and the training and test sets do not include the same writers. There are a total of 4,250 classes, but the number of classes in the training set is just 4,058. Therefore, some classes in the testing set do not appear in the training set and can not be correctly recognized.

\begin{table}
	\begin{center}
		\caption{The standard partition of the IAM dataset.}
		\label{IAM}
		\begin{tabular}{|c|c|c|}
			\hline
			Set name &  Text lines & Writers    \\
			\hline
			Train   & 6,161   & 283 \\
			\hline
			Validation1    &900   &46 \\
			\hline
			Validation2    &940  &43 \\
			\hline
			Test    &1,861  &128\\
			\hline
			Total    &9,862  &500 \\
			\hline
		\end{tabular}
	\end{center}
\end{table}

For the English handwritten text recognition, we evaluate the performance of our method on the IAM dataset. The IAM dataset contains a total of 9,862 text lines written by 500 writers. It provides one training set, one test set and two validation sets. The text lines of all data sets are mutually exclusive, thus, each writer has contributed to one set only.

\begin{figure}[!t]
	\centering
	\includegraphics[width=4.8in,height=3.5in]{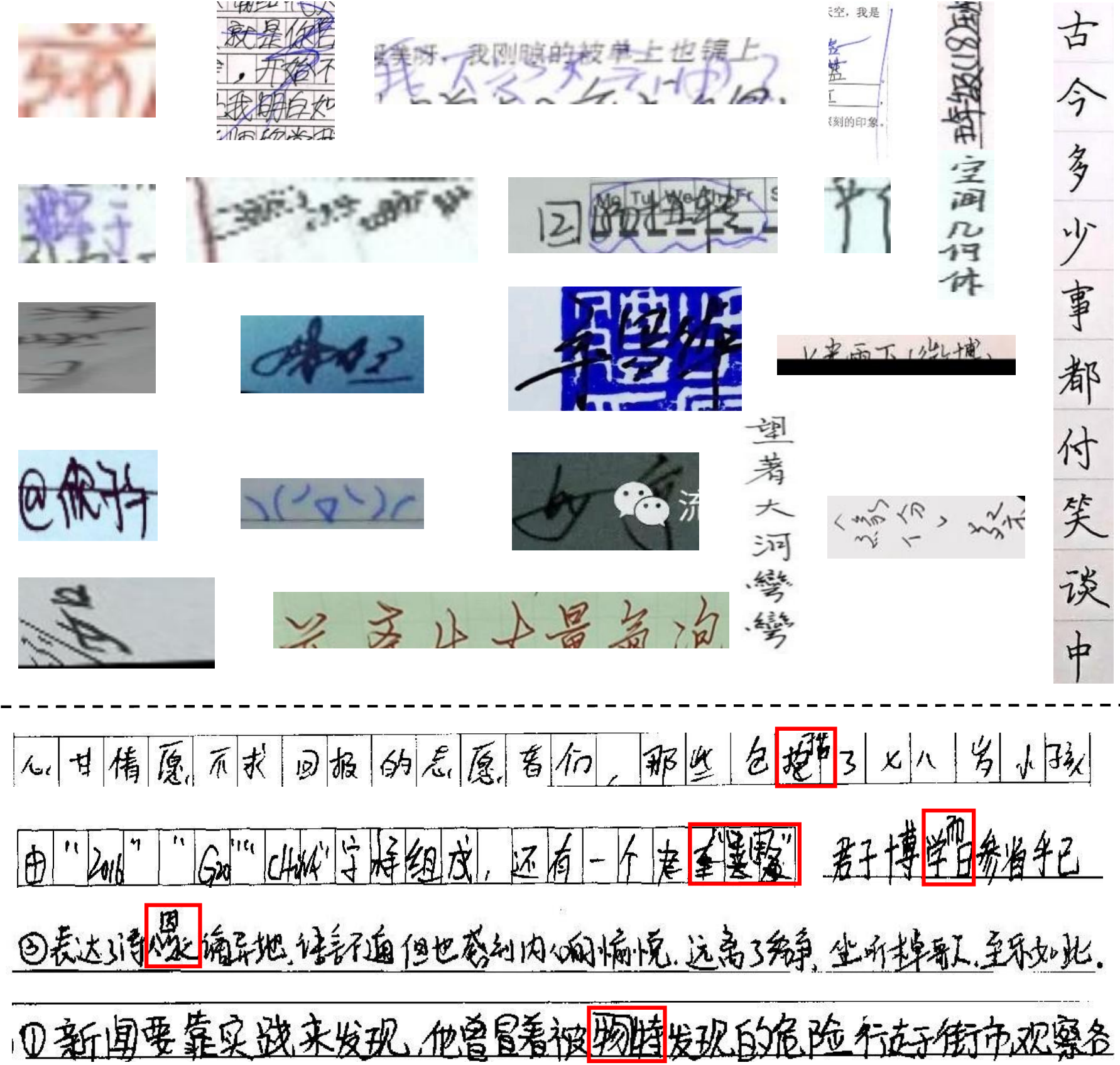}
	\caption{Some removed images in the SCUT-HCCDoc dataset are shown above the dashed line. They are very low-quality (illegible text, random graffiti, high overlap of text and background, and incomplete characters) or vertical writing text. The lower part of the dashed line shows typical removed images in the SCUT-EPT dataset. These images include abnormal structures, i.e., characters swapping and overlap, and more details can be found in \cite{wang2024leveraging}.}
	\label{abnormal}
\end{figure}

Table~\ref{Stanum} lists the number of text lines, characters and classes in both Chinese datasets. In the SCUT-EPT dataset \footnote{https://github.com/HCIILAB/SCUT-EPT\_Dataset\_Release. Prof. Lianwen Jin (eelwjin@scut.edu.cn) at South China University of Technology (SCUT) is the owner of this dataset.}, we removed 681 training text images including abnormal structures, i.e., characters swapping and overlap. Similarly, in the SCUT-HCCDoc dataset \footnote{https://github.com/HCIILAB/SCUT-HCCDoc\_Dataset\_Release. Prof. Lianwen Jin (eelwjin@scut.edu.cn) at SCUT is the owner of this dataset. Thanks to his team for building two useful datasets.}, some very low-quality (illegible text, random graffiti, high overlap of text and background, and incomplete characters) or vertical writing text images were removed and only 91,261 text instances were used in the training set. All test images in both datasets were evaluated. Table~\ref{IAM} lists the standard partition of the English dataset \footnote{https://fki.tic.heia-fr.ch/databases/iam-handwriting-database. Accessed 2021.}. All training images, validation images and test images were used in our experiments and the best model was selected for testing according to the results on the validation sets. The character accuracy rate (AR) was used to evaluate the performance of the text recognition model. The evaluation criterion is defined as follows:
\begin{equation}
  1 - \frac{{{N_\text{s}} + {N_\text{i}} + {N_\text{d}}}}{N}
\end{equation}
where $N$ is the total number of samples in the evaluation set. $N_\text{s}$ , $N_\text{i}$ and $N_\text{d}$ denote the number of substitution errors, insertion errors and deletion errors, respectively.

\subsection{Experimental Results and Analysis}
\noindent During the training stage, common data argument methods including text location shift, image blur, grey level and contrast changes, and linear transformation were used. The Adam optimization algorithm with default parameters was adopted, we adjusted the learning rate according to training steps until the model converged into a small range. At certain epochs, the learning rate was multiplied by 0.5. The deep learning platform Pytorch \cite{paszke2019pytorch} and an NVIDIA RTX A6000 with 48GB memory were used. Firstly, we conducted ablation experiments to verify the effectiveness of the proposed network.

\subsubsection{Ablation Experiments}
\noindent In this section, we examine several important factors for recognition performance. These factors include the frame resolution, the 3D attention module, the global-local context information and the multi-scale training. 

We only compare the results of different frame lengths and all sliding windows have the same height. Table~\ref{Re_res} shows the different results of different frame resolutions. We can observe that frame length 3 is the optimal configuration for all experiments, which means the corresponding window size is suitable for most characters. For a longer or shorter feature sequence, it is difficult to discriminate different regions of characters. For example, when we used a small frame length, the Chinese recognition results obviously decreased from 88.8\% to 88.06\%, 76.67\% to 75.13\%, respectively, while the English recognition performance only changed from 94.17\% to 94.11\%, which may be owing to a smaller width for most English characters. Furthermore, we can obtain consistent improvements by integrating multiple resolutions, i.e., the ARs achieve 89.16\%, 76.96\% and 94.37\%, respectively. As shown in Fig.~\ref{nn}, in the training stage, although three 3D attention blocks and global-local context blocks were used in parallel, we only retained the corresponding branch of the frame length 3 during the inference stage. 

\begin{table}
	\begin{center}
		\caption{The different results of different frame resolutions. }
		\label{Re_res}
		\begin{tabular}{|c|c|c|c|}
			\hline
			Frame length     &SCUT-HCCDoc &SCUT-EPT & IAM\\
			\hline
			2 & 88.06 & 75.13 & 94.11 \\
			\hline
			3 & 88.80  & 76.67 & 94.17  \\
			\hline
			4 & 87.21 & 76.36 & 93.62\\
			\hline
			Multi-scale & ${\bm{89.16}}$ & ${\bm{76.96}}$ & ${\bm{94.37}}$ \\
			\hline
		\end{tabular}
	\end{center}
\end{table}

Based on the optimal frame length and single-scale training, Table~\ref{Re_3d_ls} lists the results with/without the 3D attention module and the global-local context information. If we did not adopt the 3D attention module in experiments, a pooling operation was directly used to reduce the height of the last convolutional feature maps to 1. The 3D attention module can steadily improve the Chinese recognition performance. On the SCUT-HCCDoc data set, the AR is improved from 88.51\% to 88.8\%. On the SCUT-EPT data set, the AR increases from 75.00\% to 76.67\%. Although it seems that the English recognition without the 3D attention module has a higher AR (94.31\% vs. 94.17\%), the network based on the multiple-resolutions training still obtains the best AR. As the conclusions demonstrated by many previous researches, there exists an obvious gap with/without using the context information for the sequence task. The results dramatically decrease from 88.8\% to 85.36\%, 76.67\% to 72.28\% and 94.17\% to 87.0\%, respectively. It is interesting to observe that the pure CNN-based models without the 3D attention block and the context information can also achieve 85.76\%, 72.53\% and 85.93\%, respectively.
\begin{table}
	\begin{center}
		\caption{The recognition performance with/without different blocks.}
		\label{Re_3d_ls}
		\scalebox{0.93}{
		\begin{tabular}{|c|c|c|c|c|}
			\hline
			3D attention module&Global-local context information                                                  
			&SCUT-HCCDoc
			&SCUT-EPT
			&IAM
			\\
			\hline
			$\checkmark$ & $\checkmark$ & $\bm{88.80}$ & $\bm{76.67}$ & 94.17\\
			\hline
			$\checkmark$ & $\bm{\times}$ & 85.36 & 72.28 & 87.00 \\
			\hline
			$\bm{\times}$  & $\checkmark$ & 88.51 & 75.00 & $\bm{94.31}$ \\
			\hline
			$\bm{\times}$  & $\bm{\times}$ & 85.76 & 72.53 & 85.93\\
			\hline
		\end{tabular}}
	\end{center}
\end{table}

In Table~\ref{Ts}, we compare the results of different training strategies, i.e., the weights are only optimized by the CTC loss or jointly trained by the CTC loss and the CE loss. Although the Chinese decoding results of the auxiliary network are not good, the main recognition network still benefits from the joint training. The best network can achieve 89.34\%, 77.15\% and 94.41\%, respectively. Moreover, we also list the storage and the running time comparisons of different methods. In order to make a fair comparison, all experiments were evaluated on the same machine and the experimental configurations on the same dataset were consistent. In Table~\ref{Ts}, the decoding time of the CTC is defined as 1. Generally speaking, considering the auxiliary decoder with attention needed and the cyclic decoding way used during the inference stage, it is reasonable that the ED method needs more time consumption and storage. As shown in Fig~\ref{nn}, for the joint training strategy of CTC and CE, only the faster CTC method is used in the test stage, i.e., it has the same time consumption with the pure CTC training strategy.

\begin{table}
	\begin{center}
		\caption{The comparisons of accuracy rate, storage and relative speed.}
		\label{Ts}
		\begin{tabular}{|c|ccc|ccc|ccc|}
			\hline
			\multirow{2}{*}{Training strategy} & \multicolumn{3}{c|}{SCUT-HCCDoc} & \multicolumn{3}{c|}{SCUT-EPT} &
			\multicolumn{3}{c|}{IAM}    \\ \cline{2-10} 
			& \multicolumn{1}{c|}{AR} & \multicolumn{1}{c|}{Speed} & \multicolumn{1}{c|}{Storage} & \multicolumn{1}{c|}{AR} & \multicolumn{1}{c|}{Speed} & \multicolumn{1}{c|}{Storage} &
			\multicolumn{1}{c|}{AR} & \multicolumn{1}{c|}{Speed} &
			\multicolumn{1}{c|}{Storage} \\ \hline
			CTC                  & \multicolumn{1}{c|}{89.16} & \multicolumn{1}{c|}{1.0}  & 41.92MB & \multicolumn{1}{c|}{76.67} & \multicolumn{1}{c|}{1.0} & 35.62MB
			& \multicolumn{1}{c|}{94.37} & \multicolumn{1}{c|}{1.0} & 22.04MB
			\\ \hline
			CE                  & \multicolumn{1}{c|}{85.53} & \multicolumn{1}{c|}{3.10} & 46.89MB & \multicolumn{1}{c|}{72.58} & \multicolumn{1}{c|}{6.24} & 41.50MB 
			& \multicolumn{1}{c|}{93.94} & \multicolumn{1}{c|}{2.63} & 30.15MB
			\\ \hline
			CTC + CE                 & \multicolumn{1}{c|}{$\bm{89.34}$} & \multicolumn{1}{c|}{1.0}  & 41.92MB & \multicolumn{1}{c|}{${\bm{77.15}}$} & \multicolumn{1}{c|}{1.0} & 35.62MB 
			& \multicolumn{1}{c|}{$\bm{94.41}$} & \multicolumn{1}{c|}{1.0} & 22.04MB
			\\ \hline
		\end{tabular}
	\end{center}
\end{table}

\begin{table*}
	\begin{center}
		\caption{Performance comparison of our proposed method and other state-of-the-art methods. For the English handwritten recognition task, all listed results do not use any lexicons.}
		\label{Comp1}
		\begin{tabular}{|c|c|c|c|}
			\hline
			Dataset &Method     &Without external data      &With external data\\
			\hline
			\multirow{4}{*}{SCUT-HCCDoc} & Zhang et al. \cite{zhang2020scut} & 78.65 & 79.84 \\
			\cline{2-4}
			& Peng et al. \cite{peng2022recognition}& -  & $\bm{90.85}$ \\
			\cline{2-4}
			&Lin et al. \cite{lin2023building}& - & 88.10 \\
			\cline{2-4}
			&Ours                & $\bm{89.34}$  & - \\
			\hline
			\hline
			\multirow{4}{*}{SCUT-EPT} &Zhu et al. \cite{zhu2018scut} & 75.37 & 75.97 \\
			\cline{2-4}
			&Hoang et al. \cite{hoang2021lodenet}& 76.61  & $\bm{77.61}$ \\
			\cline{2-4}
			&Ngo et al. \cite{ngo2021recurrent}& 76.85 & - \\
			\cline{2-4}
			&Ours                & $\bm{77.15}$ & - \\
			\hline
			\hline
			\multirow{4}{*}{IAM} & Dutta et al.\cite{dutta2018improving} & - & 94.3  \\
			\cline{2-4}
			& Chowdhury et al. \cite{chowdhury2018efficient}& 91.9 & - \\
			\cline{2-4}
			&Ours                &  $\bm{94.41}$  & - \\
			\hline
		\end{tabular}
	\end{center}
\end{table*}

\subsubsection{Overall Comparison}
\noindent Finally, Table~\ref{Comp1} shows an overall comparison of our proposed method and other state-of-the-art methods on different test sets. Without using external data, our proposed network can make a new milestone. Compared with the best networks trained by using external data, the proposed network can also achieve comparable results on all datasets. For the listed results on the IAM dataset, the standard partition and no word lexicons were used. Please note that there exists an unofficial partition (RWTH partition) of the IAM dataset. It is widely used by many researchers, their results are not compared in this paper.

\subsubsection{Visualization Analysis}
\begin{figure}[!t]
	\centering
	\includegraphics[width=3.9in,height=2.5in]{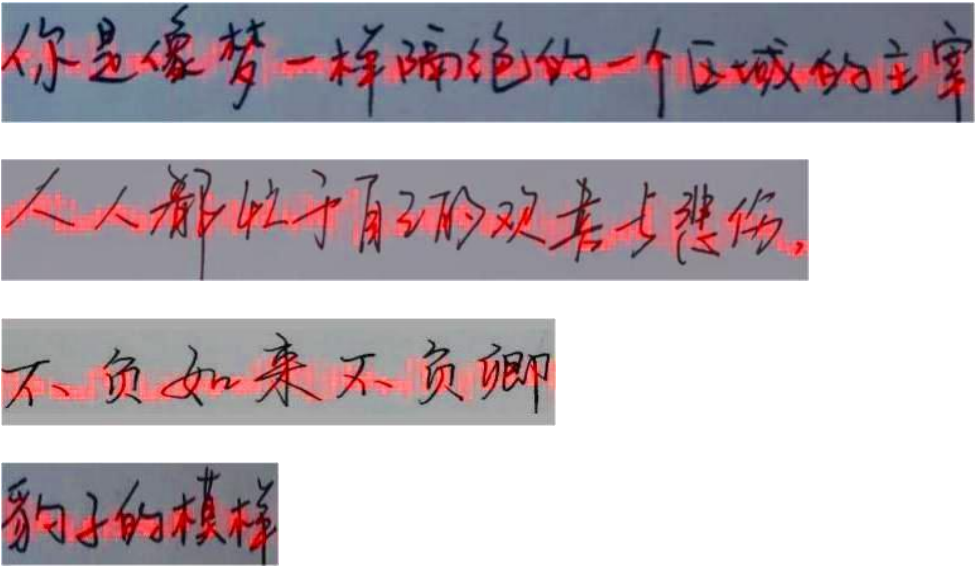}
	\caption{The typical samples of the attention visualization on the SCUT-HCCDoc images \cite{zhang2020scut}.}
	\label{vis1}
\end{figure}

\begin{figure*}[!t]
	\centering
	\includegraphics[width=4.8in,height=2.2in]{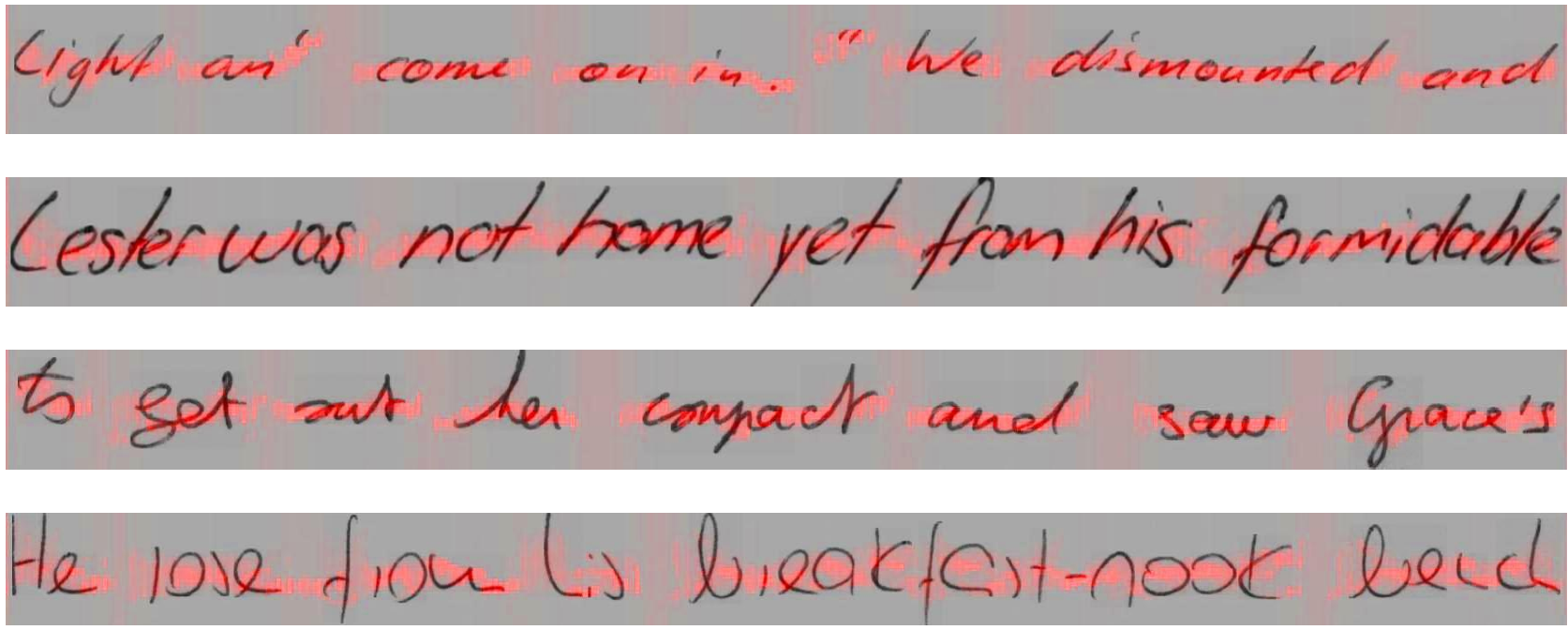}
	\caption{The visualization of the 3D attention weights on the IAM images \cite{marti2002iam}.}
	\label{vis2}
\end{figure*}

\begin{figure*}[!t]
	\centering
	\includegraphics[width=4.8in,height=2.4in]{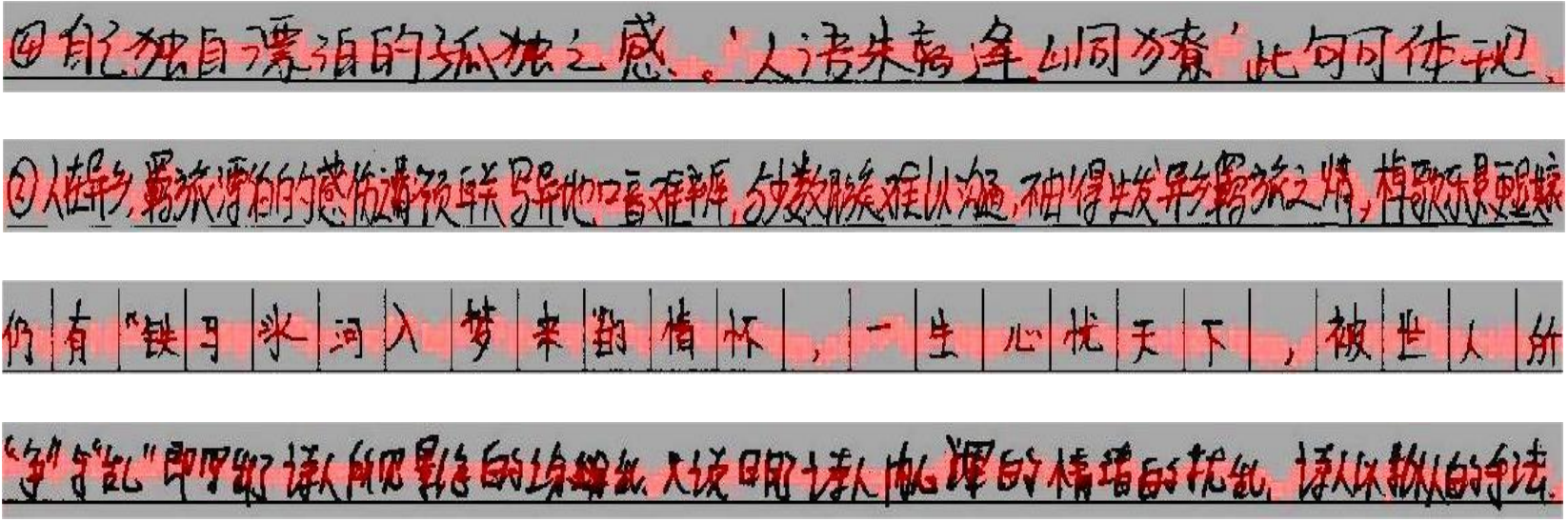}
	\caption{The typical samples of the attention visualization on the SCUT-EPT images \cite{zhu2018scut}.}
	\label{vis3}
\end{figure*}

\noindent In Fig.~\ref{vis1}, Fig.~\ref{vis2} and Fig.~\ref{vis3}, the attention weights in the 3D block are shown on the original images. It is interesting to observe that even for long-text images and some small symbols, most attention weights (red parts) are still located on the handwriting characters, which demonstrates the proposed attention mechanism is reasonable.



\section{Discussion and Conclusion}
\label{DisCon}
\noindent From the above experiments, we can observe that the proposed network has achieved good results on multiple handwritten text datasets. One important factor is that the proposed network can absorb the advantages of three typical segmentation-free sequence recognition algorithms. Compared with the HMM, which relies on artificially designed features and the generative model (GMM-HMM) for character modeling, we do not need to perform complex character modeling and inference. However, we have integrated the state-level feature extraction method based on sliding windows in HMM into the intermediate process of the network, and ingeniously extracted multi-scale block features for the training of the network, which potentially increases the training data and enables the network to better handle characters of different sizes. Compared with the encoder-decoder method based on the attention mechanism, which is prone to attention drift when applied to long handwritten text recognition, the proposed 3D attention mechanism is conducted on the sliding window to generate better local visual features. Whether from intuition or from our experimental results, the proposed 3D attention can better take into account the 2D natural properties of characters. Without too many tricks, we use the most popular neural units to implement the 3D attention. The extracted sequence visual features are used to obtain global and local context features through the self-attention mechanism and the LSTMs, respectively. Obviously, the representation after combining visual features and contextual features is suitable for the text recognition task. Finally, by cascading CTC loss and CE loss, the whole network can be optimized well. As shown in Fig~\ref{nn}, in this paper, during the test stage, we only use one of the network branches and only use the faster CTC method, without the cyclic decoding way of common ED methods.

Overall, main canonical neural units including attention mechanisms, fully-connected layers, recurrent units and convolutional layers are ingeniously organized into a network. The network weights are effectively optimized by the multi-scale training. Experimental results show that the proposed method can achieve comparable results with the state-of-the-art methods. Besides, the visualization analysis demonstrates the proposed attention mechanism is reasonable. For future work, we will investigate large vision and language models for document analysis.


\section*{Acknowledgments}
This work was supported by the National Natural Science Foundation of China under Grant No. 62106031. The author would like to thank Professor Jun Du of the University of Science and Technology of China (USTC). When the author was a Ph.D. candidate at USTC (September 2015 to June 2020), we successfully applied the HMM and the deep neural network to the text recognition task. The supervisor (Prof. Jun Du) always said, can we combine three mainly segmentation-free methods? Although the HMM is not explicitly used in this paper, the question inspired the author to delve into this problem while self-isolated at home. The author also would like to thank Yan Wang at USTC for applying the code of extracting text lines from original SCUT-HCCDoc documents.

%
%
%
%




\bibliographystyle{unsrt}
\bibliography{ref.bib}

\end{document}